\documentclass[10pt,twocolumn,letterpaper]{article}

\usepackage{cvpr}
\usepackage{times}
\usepackage{epsfig}
\usepackage{graphicx}
\usepackage{amsmath}
\usepackage{amssymb}
\usepackage{bbm}
\usepackage{multirow}
\usepackage{float}
\usepackage[pagebackref=true,breaklinks=true,colorlinks,bookmarks=false]{hyperref}

\cvprfinalcopy % *** Uncomment this line for the final submission

\begin{document}

%%%%%%%%% TITLE
\title{Stronger Baseline for Person Re-Identification}

\author{Fengliang Qi, Bo Yan, Leilei Cao, Hongbin Wang\\
Ant Group\\
}

\maketitle
%\maketitle
%\thispagestyle{empty}

%%%%%%%%% ABSTRACT
\begin{abstract}
   Person re-identification (re-ID) aims to identify the same person of interest across non-overlapping capturing cameras, which plays an important role in visual surveillance applications and computer vision research areas. Fitting a robust appearance-based representation extractor with limited collected training data is crucial for person re-ID due to the high expanse of annotating the identity of unlabeled data. In this work, we propose a Stronger Baseline for person re-ID, an enhancement version of the current prevailing method, namely, Strong Baseline, with tiny modifications but a faster convergence rate and higher recognition performance. With the aid of Stronger Baseline, we obtained the third place (i.e., 0.94 in mAP) in 2021 VIPriors Re-identification Challenge without the auxiliary of ImageNet-based pre-trained parameter initialization and any extra supplemental dataset.
\end{abstract}

\section{Introduction}\label{sec1}
Person re-identification (re-ID) has become a core, and widely used technique in visual surveillance applications and computer vision research areas \cite{sun2019mvp, yan2019hornet, liu2019spatially}. 
It aims at locating and recognizing a person of interest across multiple non-overlapping cameras in various spots \cite{zhang2019scan, subramaniam2019co, fu2019sta}.

\begin{figure}[!htb]
	\centering
	\includegraphics[width=0.9\columnwidth]{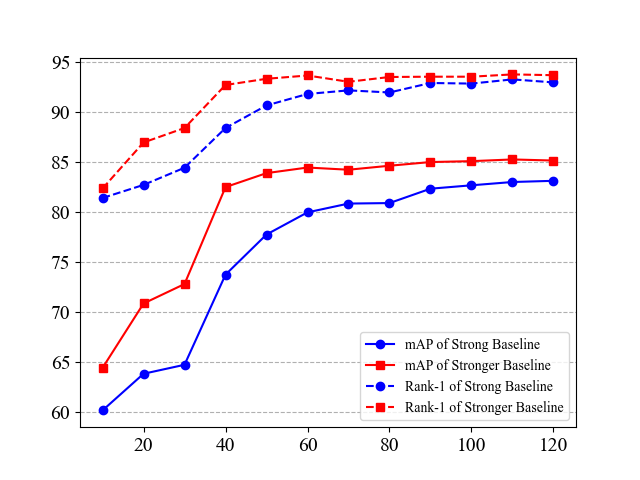}
%  	\vspace{-0.1cm} 
	\caption{	
		\textbf{Convergence rate and performance are both enhanced considerably on Market1501 via the proposed Stronger Baseline.}
	}
\label{fig1}
% 	\vspace{-0.5cm}
\end{figure} 

The main challenge for the person-reID task is that the training dataset can be limited comparing with the conventional benchmark for classification task (e.g., ImageNet), making the risk of over-fitting increasing, which causes to the deterioration of the final generalization performance. To alleviate this dilemma, most of the current work focus on two main technical routines: 1) Designing customized and lightweight neutral network structure (e.g., OSNet); 2) Introducing discriminative margin-based loss function (e.g., Triplet loss) which has been widely used in metric learning and related research field (e.g., face verification).

In this work, we follow the second technical routine and propose the \textbf{Stronger Baseline} by modifying the Strong Baseline proposed by \cite{Luo2019CVPRWorkshops} with tiny modification but faster convergence rate and higher recognition performance as shown in Fig\ref{fig1}. In contrast to the argument claimed in \cite{Luo2019CVPRWorkshops}, we argue that the \textbf{BNNeck} is not the critical factor in alleviating the conflict between the triplet loss and cross-entropy loss minimization process. 
Instead, we claim the BNNeck is simply a standardization procedure. When combined with a softmax classifier, whose output follows the multinomial distribution, one of the exponential family distribution, the loss landspace can be smoother. The gradient update direction can be more stable, making the optimization algorithm more possible to arrive at the optimal global solution.

\begin{figure*}[!htb]
	\centering
	\includegraphics[width=1.8\columnwidth]{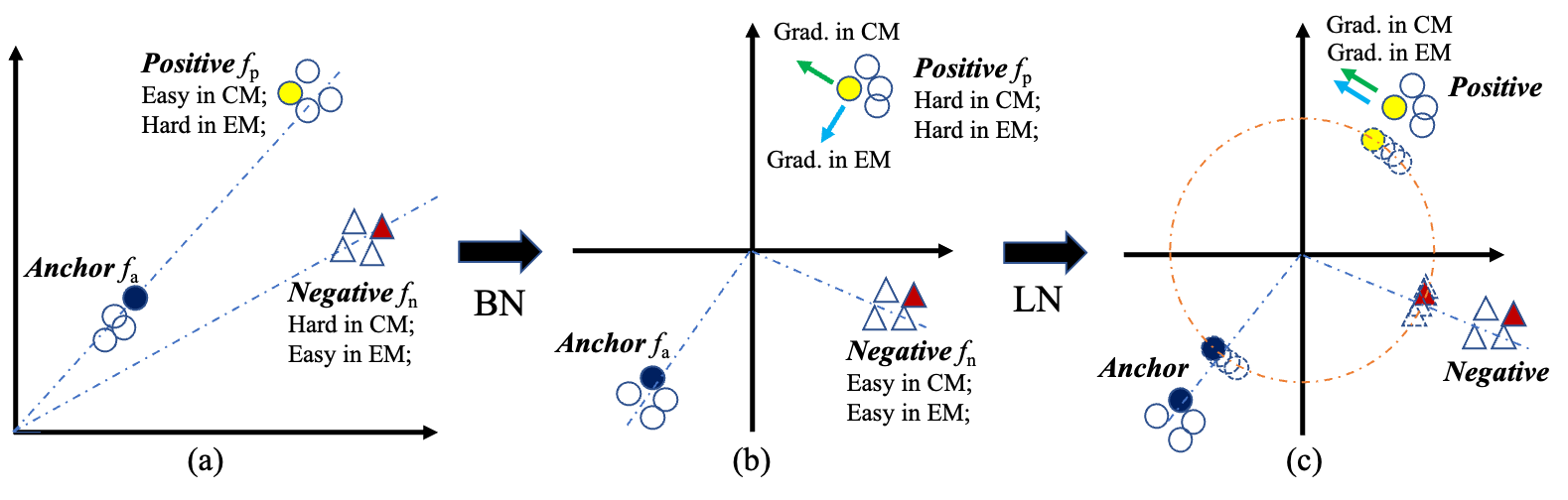}
% 	\vspace{-0.3cm} 
	\caption{	
		\textbf{Schematic illustration of the optimization conflict in Cosine and Euclidean metric space.} (a) The hardness of samples is identified inconsistently in Cosine and Euclidean metric space. (b) BN module can alleviate the first inconsistency in identifying the hardness of samples. (c) LN operation can alleviate the second inconsistency in gradient updating direction. Shape refers to the identity of the samples, and filling color refers to the role in metric pairs (e.g., anchor), Grad., BN, LN, CM, and EM refer to the Gradient, Batch-Normalization, L2 Normalization, Cosine Metric, and  Euclidean Metric, respectively.
	}
\label{fig2}
% 	\vspace{-0.5cm}
\end{figure*}

Furthermore, we attribute the occurrence of optimization conflict between cross-entropy loss and triplet loss to two main aspects: 1) inconsistency in identifying the hard samplers during the hard mining process and 2) inconsistency in gradient update direction during the objective minimizing process in two metric space (i.e., Cosine Metric Space and Euclidean Metric Space).
As illustrated in Fig.\ref{fig2}(a), compared to the anchor sample $f_a$, the positive sample $f_p$ is easy in Cosine Metric Space since it lies in the similar radial direction with anchor sample $f_a$; however, when it comes to the Euclidean Metric Space, the positive sample $f_p$ should be treated as a hard sample since its significant Euclidean distance from anchor sample. The inconsistency in identifying the hardness of the positive sample holds for the negative sample (e.g., $f_n$ in Fig.\ref{fig2}(a)) in a similar principle as well. To alleviate this conflict, we use Batch-Normalization  (BN) module to standardize the distribution of the samplers as shown in Fig.\ref{fig2}(b) and make the identification of samples' hardness consistent in both Cosine and  Euclidean metric space. 

As for the second inconsistency in gradient update direction, the gradient for the positive sample $f_p$ in Euclidean Metric Space is parallel to the link between $f_a$ and $f_p$. However, in Cosine Metric Space, the gradient update direction is parallel to the tangent direction, confusing the final comprehensive updating direction and deteriorating the optimization process. To mitigate this confusion, we use the L2 Normalization operation for the feature before triplet loss estimation. Thus the triplet loss can optimize the distance in the same cosine metric space with cross-entropy loss as illustrated in Fig.\ref{fig2}(c).

\begin{figure*}[!htb]
%\begin{figure}[H]
% 	\vspace{-0.2cm}
	\begin{center}
	\includegraphics[width=1.5\columnwidth]{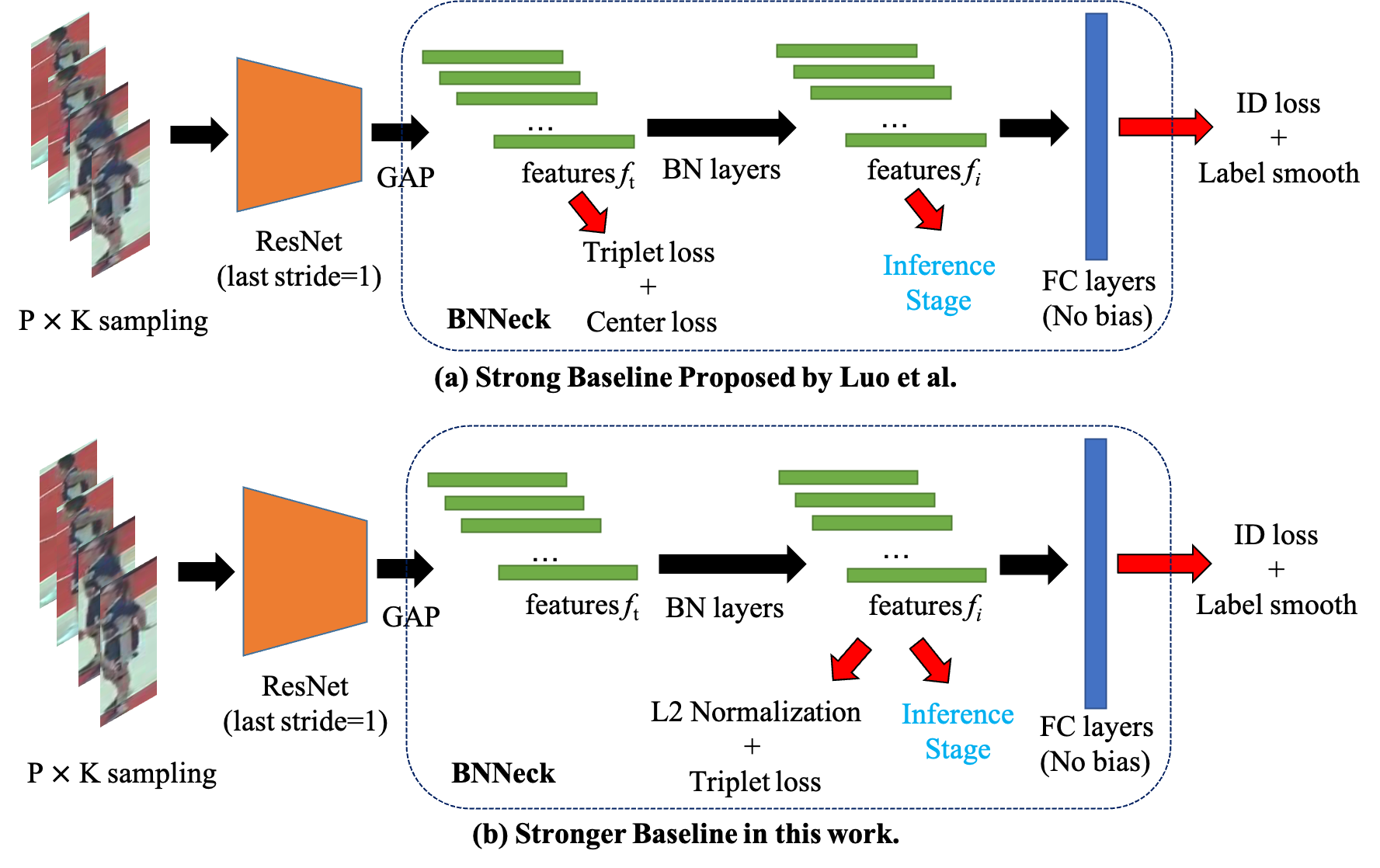}
% 	\vspace{-0.2cm}
    \end{center}
	\caption{\textbf{The overview of the pipeline of Stronger Baseline}. (a) The primary pipeline of Strong Baseline. (b) The enhanced version of Strong Baseline, namely, Stronger Baseline.}
	\label{fig3} 
 	\vspace{-0.5cm}
\end{figure*}

Based on the analysis for alleviating the conflict in optimizing the cross-entropy loss and triplet loss, we propose our \textbf{Stronger Baseline} with tiny modification on the Strong Baseline proposed in \cite{Luo2019CVPRWorkshops}. As illustrated in Fig.\ref{fig3}(b), the feature $f_t$ extracted from the backbone network (e.g., ResNet) is processed by subsequent Batch-Normalization module to generate the feature $f_i$ for cross-entropy loss calculation in the training stage and similarity evaluation in inference stage, which is the same as the pipeline of Strong baseline in Fig.\ref{fig3}(a). Different from Strong Baseline, the \textbf{Stronger Baseline} calculate the triplet loss on the L2 normalization version of feature $f_i$ instead of feature $f_t$. We discard the center loss for its limited influence on the final performance\cite{Luo2019CVPRWorkshops}. 

The main contribution of this work can be summarized in the following folds:
	
1. We observe two kinds of inconsistency when simultaneously optimizing the Cosine and Euclidean metric space, inducing the conflict in minimizing the cross-entropy and triplet loss.

2. We propose the \textbf{Stronger Baseline} based on analysis of how to alleviate the inconsistency during optimization by ameliorating the Strong Baseline with the limited modification.

3. We verify the superiority of our proposed method on the Market1501 and SynergySports benchmark.

\section{Related Works}\label{sec2}
	\subsection{Person Re-identification}\label{sec21}
Gheissari et al. \cite{gheissari2006person} first defined the person re-ID as a specific computer vision task. 
Most of the traditional methods focused on feature extraction via handcraft design before the emergence of deep learning\cite{kviatkovsky2012color,farenzena2010person,gray2008viewpoint,matsukawa2016hierarchical,cheng2011custom}. 
With the rise of deep learning in recent years, convolutional neural networks (CNN) based feature representation methods have become the mainstream for image-based person re-ID \cite{li2018harmonious,yan2019hornet,sun2018beyond,xiao2016learning}.
According to Zheng et al. \cite{zheng2016person}, most of the pre-existing image-based works concentrated on discriminative learning and metric learning. Discriminative learning \cite{li2018harmonious,chang2018multi,guo2018efficient,liu2017end} aims at getting representative features for identity classification. While metric learning \cite{sun2017svdnet,zheng2017person} learns to project extracted features from different cameras and views into a common feature representation subspace.

\subsection{Distance Metric Learning}\label{sec23}	
The goals of the training and testing are slightly different in person re-ID tasks. The training process mainly focuses on classification or metric learning, while the testing process is a retrieval problem. 
Most existing supervised person re-ID methods apply identification loss for identity classification (e.g., cross-entropy loss)\cite{sun2017svdnet} and verification loss for metric learning (e.g., triplet loss)\cite{hermans2017defense}. Triplet loss \cite{schroff2015facenet} is designed initially for face recognition problem and has been regarded as a commonly used method in the retrieval related tasks, especially in person re-ID. In the triplets, an anchor, a positive sample, and a negative sample are included.
Since triplet loss is calculated by two randomly sampled person identification, it is difficult to ensure that the distance between the anchor and positive samples are smaller than the distance between the anchor and negative examples in the whole training dataset\cite{Luo2019CVPRWorkshops}.
Quadruplet loss\cite{Chen2017CVPR} is an improved version of triplet loss, which contains two different negative samples to learn a larger inter-class distance and a smaller intra-class distance compared with the triplet loss\cite{jiang2020weighted}.

\section{Our Method}\label{sec3}

\subsection{Holistic Pipeline}\label{sec31}
In our experiments, the Stronger Baseline is applied as our holistic pipeline to realize the representation extracting and loss objective calculation. We use ResNet50 as a backbone for controlled experiments and \textit{OSNet} \textbf{without ImageNet pretrain} for final result submission. The influence of the Backbone is empirically limited to the yielded performance, and the gain of the performance mostly attributes to the proposed \textbf{Stronger Baseline}. Both the BNNeck and FC layers are initialized through Kaiming initialization proposed in\cite{he2015delving}.  Each image in a  training batch is prepossessed and resized to 256$\times$128 in pixels. In each batch, we sample 8 identities, each with 4 images.
\subsection{Data Augmentation}\label{sec32}
We utilize the following data augmentation for enhancing the generalization performance: \textit{Random Horizontal Flip} with probability 0.5, \textit{Random Erasing} with probability 0.5,  \textit{Color Jitter} with probability 0.5, and \textit{AutoAugmentation}\cite{DBLP:journals/corr/abs-1805-09501}. We are inconclusive in which augmentation strategy is beneficial to the final performance at most. We solely adopted the first two augmentation strategies while conducting the controlled experiment. 
\subsection{Post-processing}\label{sec33}
Person re-ID can also be regarded as a retrieval problem. While presenting our final results, we utilize two common post-processing strategies, i.e., Query Expansion and ReRank, whose hyperparameters follow the default setting in FastReID.

\subsection{Optimization Strategy}\label{sec34}
For controlled trials on Market1501, Adam\cite{kingma2014adam} optimizer is selected with an initial learning rate of $3.5\times10^{-4}$. The commonly adopted warm-up strategy\cite{fan2019spherereid} is applied to bootstrap the network for better performance. In practice, the network is optimized for 120 epochs. We spend $10$ epochs linearly increasing the learning rate from $3.5\times10^{-6}$ to $3.5\times10^{-4}$, and it then decays at the 30th and 55th epoch.
We discard the warm-up stage for final submission and use SGD optimizer instead due to the exclusion of pretrain. The network is optimized for 350 epochs, and the init learning rate of 0.065 decays at the 150th, 225th and 300th epoch by 0.1.

\section{Experiments}\label{sec4}
We conduct two main experiments in this work, i.e., Controlled Experiment on the public benchmark (i.e., Market1501) and Submitted Experiment on SynergySports. The former aims to justify the superiority of the proposed Stronger Baseline, and the latter aims to obtain a better result with extra tricks (e.g., Complex Data Augmentation and Post-Processing Strategy).

\subsection{Datasets and Evaluation Protocol}\label{sec41}
\textbf{Market-1501} dataset is featured by 1,501 IDs, 19,732 gallery images and 12,936 training images captured by 6 cameras. Market-1501 are produced by the DPM detector. The Cumulative Matching Characteristics (CMC) curve is used for performance evaluation, which encodes the possibility that the query person is found within the top n ranks in the rank list. We also employ the mean Average Precision (mAP), which considers the retrieval process's precision and recall.
The evaluation toolbox provided by the Market-1501 authors is used.

\textbf{SynergySports} generates from short sequences of basketball games, and each sequence is composed of 20 frames. For the validation and test sets, the query images are persons taken at the first frame, while the gallery images are identities taken from the 2nd to the last frame. There are 436 IDs, 8569 images in training split, and 50 IDs, 960 images in validation split.

\subsection{Empirical Results}\label{sec42}
As illustrated in Fig\ref{fig1}, the Stronger Baseline can enhance the convergence rate and performance simultaneously. The detailed results are listed in Table\ref{tab1}. The mAP of Strong Baseline is improved by 2.03, which is a considerable enhancement considering the tiny modification and no extra introduced overhead. We further evaluate our method on the validation set of SynergySports; the results are listed in Table\ref{tab2}, the mAP achieves the predictable improvement by 1.89.

As for the Submitted Experiment, we switch to the OSNet, a lightweight backone customized for person re-ID, and introduce the complex data augmentation and post-processing strategy for better generalization performance. The results on the validation and test set are listed in Table\ref{tab3}. 

\begin{table}[htb]
    
    %\vspace{-1em}
    \small
    \renewcommand{\arraystretch}{1.0}
	\centering
	%	\vspace{-0.2cm}
	%\resizebox{\linewidth}{!}{
		\begin{tabular}{c|c|c}
	\hline
	%\toprule
	{\bf Protocol}&
	%\multicolumn{1}{c|}{\bf Triplet}&\multicolumn{3}{c}{\bf PhD}\cr\cline{3-5}
	{\bf Strong Baseline}&{\bf Stronger Baseline}\cr\cline{1-3}
	%\midrule
	\hline
	mAP  &83.13 &85.16 (+2.03) \cr
	Rank-1  &92.99 &93.71 (+0.72) \cr
	\hline
	%\bottomrule
    \end{tabular}
	%}
	%\vspace{-0.2cm}
    
	%\vspace{0.2cm}
	\caption{The mAP and rank-1 comparison result of Strong Baseline and Stronger Baseline on Market1501. We use ResNet50 as backbone and ignore the complex data augmentation (i.e., Color Jitter and AutoAug) and Post-Processing here.}
    \label{tab1}
\end{table}

\begin{table}[htb]
    
    %\vspace{-1em}
    \small
    \renewcommand{\arraystretch}{1.0}
	\centering
	%	\vspace{-0.2cm}
	%\resizebox{\linewidth}{!}{
		\begin{tabular}{c|c|c}
	\hline
	%\toprule
	{\bf Protocol}&
	{\bf Strong Baseline}&{\bf Stronger Baseline}\cr\cline{1-3}
	%\midrule
	\hline
	mAP  &92.74 &94.63 (+1.89) \cr
	\hline
	%\bottomrule
    \end{tabular}
	%}
	%\vspace{-0.2cm}
    
	%\vspace{0.2cm}
	\caption{ The mAP comparison result of Strong Baseline and Stronger Baseline on SynergySports (Validation Set). We use ResNet50 as backbone and ignore the complex data augmentation (i.e., Color Jitter and AutoAug) and Post-Processing here.}
	\label{tab2}
\end{table}

\begin{table}[!htb]
    %\vspace{-1em}
    \small
    \renewcommand{\arraystretch}{1.0}
	\centering
	%	\vspace{-0.2cm}
	%\resizebox{\linewidth}{!}{
		\begin{tabular}{c|c|c}
	\hline
	%\toprule
	{\bf Protocol}&
	{\bf Validation Set}&{\bf Test Set}\cr\cline{1-3}
	%\midrule
	\hline
	mAP  &95.17 &94.19 \cr
	\hline
	%\bottomrule
    \end{tabular}
	%}
	%\vspace{-0.2cm}
    \caption{The mAP result on SynergySports. We use OSNet1x0 as backbone and introduce the complex data augmentation (i.e., Color Jitter and AutoAug) and Post-Processing here.}
    \label{tab3}
	%\vspace{0.2cm}
\end{table}
%	-------------------------------
\section{Conclusions}
In this paper, we concluded two kinds of inconsistency while optimizing the Cosine and Euclidean metric space simultaneously, i.e., hardness identification inconsistency and gradient update inconsistency, causing the conflict between minimizing the cross-entropy loss and triplet loss. 
To alleviate the inconsistency and ameliorate the optimization process, we proposed the Stronger Baseline with tiny modifications on the Strong Baseline but a faster convergence rate and higher evaluation performance. With the aid of Stronger Baseline, we obtain a third place in the 2021 VIPriors Re-identification Challenge without the auxiliary of ImageNet-based pre-train parameter initialization and any extra supplemental dataset.
{\small
\bibliographystyle{ieee_fullname}
\bibliography{stronger}
}

\end{document}